\documentclass[10pt,twocolumn,letterpaper]{article}

\usepackage{cvpr}              
\usepackage{marvosym}

\usepackage{stfloats}
\usepackage[dvipsnames]{xcolor}
\usepackage{multirow}
\usepackage{array}
\usepackage[ruled]{algorithm2e}
\usepackage{colortbl}  
\usepackage{makecell}
\SetAlgoCaptionSeparator{}

\newcommand{\methodName}{\textbf{\textit{AllSpark}}}{}
\newcommand{\SM}{Semantic Memory}{}
\newcommand{\CSG}{Channel Semantic Grouping}{}

\definecolor{cvprblue}{rgb}{0.21,0.49,0.74}
\usepackage[pagebackref,breaklinks,colorlinks,citecolor=cvprblue]{hyperref}

\makeatletter
\def\customsymbol#1{
    \ifcase\number\value{#1}
        \or *
        \or \Letter
        \or 1
    \else\@ctrerr
    \fi
}
\makeatother

\def\paperID{7007} 
\def\confName{CVPR}
\def\confYear{2024}

\title{\textit{AllSpark}: Reborn Labeled Features from Unlabeled in Transformer for Semi-Supervised Semantic Segmentation}

\author{Haonan Wang\footnotemark[1], \  Qixiang Zhang\footnotemark[1], \ Yi Li, Xiaomeng Li\textsuperscript{\Letter}\\
The Hong Kong University of Science and Technology\\
{\tt\small \{hwanggr, qzhangcq, ylini\}@connect.ust.hk, eexmli@ust.hk}
}

\begin{document}

\maketitle

\setcounter{footnote}{1}
\renewcommand{\thefootnote}{\customsymbol{footnote}}
\footnotetext[1]{Equal contribution.} 
\setcounter{footnote}{2}
\renewcommand{\thefootnote}{\customsymbol{footnote}}
\footnotetext[2]{Corresponding author.} 

\begin{abstract}

Semi-supervised semantic segmentation (SSSS) has been proposed to alleviate the burden of time-consuming pixel-level manual labeling, which leverages limited labeled data along with larger amounts of unlabeled data. 
Current state-of-the-art methods train the labeled data with ground truths and unlabeled data with pseudo labels. However, the two training flows are separate, which allows labeled data to dominate the training process, resulting in low-quality pseudo labels and, consequently, sub-optimal results.
To alleviate this issue, we present \methodName{}\footnote{The \textit{AllSpark} is a powerful Cybertronian artifact in the film of \textit{Transformers: Revenge of the Fallen}, which can be used to reborn the Transformers. It aligns well with our core idea.}, which reborns the labeled features from unlabeled ones with the channel-wise cross-attention mechanism. We further introduce a \SM{} along with a \CSG{} strategy to ensure that unlabeled features adequately represent labeled features.
The \methodName{} shed new light on the architecture level designs of SSSS rather than framework level, which avoids increasingly complicated training pipeline designs. 
It can also be regarded as a flexible bottleneck module that can be seamlessly integrated into a general transformer-based segmentation model. 
The proposed \methodName{} outperforms existing methods across all evaluation protocols on Pascal, Cityscapes and COCO benchmarks without bells-and-whistles. Code and model weights are available at: \href{https://github.com/xmed-lab/AllSpark}{\textit{\texttt{https://github.com/xmed-lab/AllSpark}}}.

\end{abstract}    
\begin{figure}[t]
    \centering
    \includegraphics[width=\linewidth]{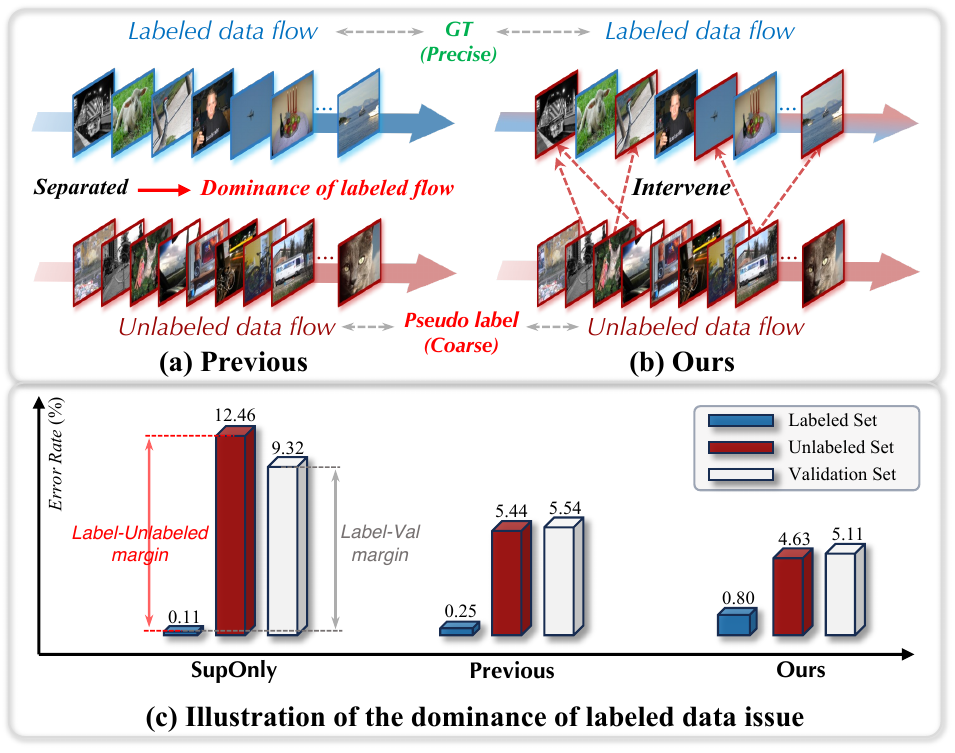}
    \caption{(a)(b) Comparison between the training data flows of previous methods and ours. Previous methods separate the labeled and unlabeled data training flows thus leading to dominance of the labeled data. (c) Dominance of labeled data issue of the previous method (\textit{e.g.}, UniMatch~\cite{UniMatch}). The \textcolor{red}{red margin} denotes how the labeled data overwhelm the unlabeled; Larger \textcolor{gray}{gray margin} indicates the model may over-fit to labeled data.
    }
    \label{fig:comparison}
\end{figure}

\section{Introduction}\label{sec:intro}

Semantic segmentation is a dense prediction task that aims to enable deep learning models to have class awareness~\cite{seg_long2015fcn,seg_ronneberger2015unet,seg_chen2018deeplabv3,vit_xie2021segformer}.
However, the success of such approaches heavily relies on labor-intensive manual labeling.
Semi-Supervised Semantic Segmentation (SSSS) methods~\cite{MT,PseudoSeg,CPS,U2PL,UniMatch,AND,DHC}, which are built upon the principles of semi-supervised learning~\cite{ssl_lee2013pseudo_label, ssl_berthelot2019mixmatch, ssl_sohn2020fixmatch, ssl_zhang2021flexmatch}, are well-studied to train models using a limited amount of labeled data in conjunction with a larger amount of unlabeled data.


Most state-of-the-art SSSS methods~\cite{UniMatch,CFCG,LogicDiag,ESL,DLG} are based on the pseudo labeling~\cite{ssl_lee2013pseudo_label,ssl_sohn2020fixmatch} scheme, where the unlabeled data is assigned pseudo labels based on model predictions. 
The model is then iteratively trained using these pseudo labeled data as if they were labeled examples. 
Based on the training flows of the labeled and unlabeled data, the current state-of-the-arts can be mainly divided into two categories: (1) \textit{successive}, the most representative one is the teacher-student-based framework~\cite{MT,U2PL,GTASeg,SemiCVT,AugSeg,DGCL}, where a student model is first trained with labeled data, then a teacher model updated by the exponential moving averaging (EMA) of the student is used to generate the pseudo labels; (2) \textit{parallel}, represented by CPS (Cross Pseudo Supervision)~\cite{CPS,UniMatch,CCVC,DLG}, where the labeled data are concatenated with the unlabeled data within a mini-batch, and the model optimize the two flows simultaneously through consistency regularization.


One key characteristic of these state-of-the-art methods is the separation of training flows for labeled and unlabeled data, as shown in Figure~\ref{fig:comparison}(a). The separation of training flows allows labeled data to dominate the training process since they have labels; see Figure~\ref{fig:comparison}(c). However, these approaches have a common significant drawback: the model used to generate pseudo labels for unlabeled images is heavily influenced by a limited set of labeled samples, resulting in low-quality pseudo labels and, consequently, sub-optimal results. Unfortunately, previous research has overlooked this crucial observation.

\begin{figure}[t]
    \centering
    \includegraphics[width=0.95\linewidth]{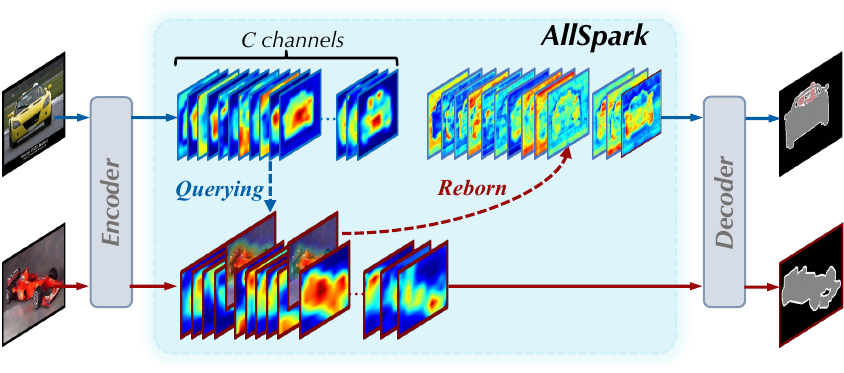}
    \caption{Illustration of the core idea of \methodName{}, which leverages the unlabeled features to reborn the labeled ones. The regenerated labeled features exhibit a high level of precision, yet they also possess diversity compared to the original features. 
    }
    \label{fig:core}
\end{figure}

\begin{figure*}[!ht]
    \centering
    \includegraphics[width=\linewidth]{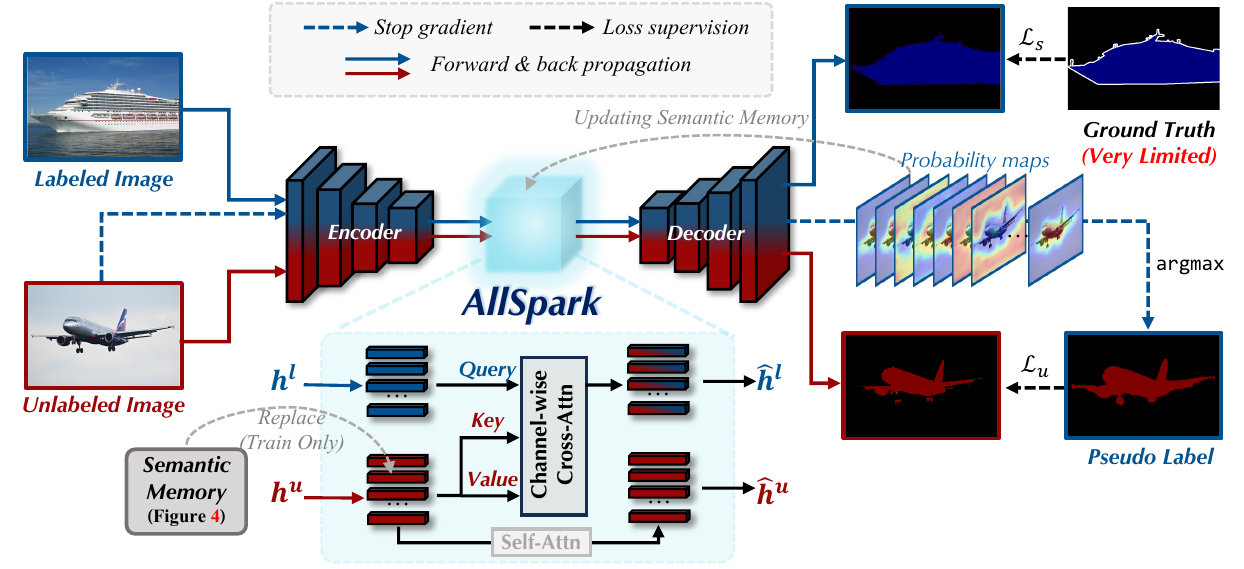}
    \caption{Illustration of the proposed \methodName{}, which can be regarded as a flexible bottleneck plugged in the middle of a general segmentation model. 
    In the training stage, the unlabeled features are replaced by the Semantic Memory (\S~\ref{sec:memory_bank} \& Figure~\ref{fig:memory_bank} bottom). Moreover, the probability maps is used for \CSG{} strategy (\S~\ref{sec:grouping} \& Figure~\ref{fig:memory_bank} top). In the inference stage, the cross-attention is degraded to self-attention with the inputs as the hidden features of the test images.}
    \label{fig:framework}
\end{figure*}

To this end, we present a novel approach: rather than training directly on the labeled samples, we leverage the unlabeled ones to generate labeled features (as depicted in Figure~\ref{fig:comparison}(b)).
The underlying rationale is that the labeled data is far less than the unlabeled data, by reborning labeled features from unlabeled data, we introduce diversity into the labeled flow, creating a more challenging learning environment. Instead of relying solely on self-reconstruction, the labeled branch learns to extract valuable channels from the unlabeled features.
The generic semantic information within channels provides feasibility to this idea, for example, as illustrated in Figure~\ref{fig:framework}, the labeled sample of a ``boat'' and the unlabeled sample of ``airplanes'' exhibit a similar metal material texture.

The subsequent step is to determine an effective method to reborn the labeled data. Fortunately, the Cross-Attention mechanism~\cite{vaswani2017attention}, initially utilized in the transformer decoder, provides a suitable approach as it is designed to reconstruct a target sequence using the source sequence.
Therefore, we develop our method based on the concept of cross-attention, adapting it to operate in a channel-wise manner~\cite{wang2022uctransnet,ding2022davit} in order to extract channel-wise semantic information. This method, named \methodName{}, utilizes the labeled data features as queries and the unlabeled data features as keys and values, as depicted in Figure~\ref{fig:core}.
Specifically, \methodName{} calculates a similarity matrix between each channel of the labeled features and the unlabeled features. The unlabeled channels with the highest similarities are then emphasized to reconstruct the labeled features.
However, a critical challenge in implementing this idea is the limited unlabeled features within a mini-batch, which may not be sufficient to reconstruct the correct labeled features.
To address this, we introduce a \SM{} (S-Mem) that gathers channels from previous unlabeled features. S-Mem are updated iteratively with a proposed channel-wise semantic grouping strategy which first groups together channels that represent the same semantics and then enqueue them into according class slot of S-Mem.

Furthermore, naively plugging \methodName{} into a CNN-based backbone is not applicable, due to the smaller reception field of the convolution operation, as shown in Figure~\ref{fig:CNNvsViT}, compared with that of the Vision Transformers (ViTs)~\cite{dosovitskiy2021vit}, which have made remarkable progress in semantic segmentation~\cite{vit_xie2021segformer,vit_wang2021pvt,vit_strudel2021segmenter,vit_cheng2021maskformer,vit_Jain_2023_oneformer,vit_jain2023semask} with the well-known property of capturing long-range dependencies. Thus, we focus on building a pure-transformer-based SSSS framework.

Our contributions can be summarized in three folds:
\begin{itemize}
    \item We notice the separate training approach in current SSSS frameworks will cause the dominance of labeled data and thus lead to sub-optimal solutions. 
    \item We propose \methodName{} module to address the  labeled data dominance issue and build a SOTA pure-transformer-based SSSS framework, which can further be benefit for efficiently training the foundation models, \textit{e.g.}, SAM~\cite{kirillov2023SAM}, with less labeled data.
    \item Extensive experiments have been conducted to validate the effectiveness of our proposed \methodName{}.
    The results of these experiments have demonstrated solid performance gains across three widely recognized benchmarks: PASCAL VOC 2012~\cite{everingham2015pascal}, Cityscapes~\cite{cordts2016cityscapes}, and COCO~\cite{lin2014coco}. 
\end{itemize}


\section{Related Work}

\subsection{Semantic Segmentation}
Semantic segmentation is a dense prediction tasks, and has been significantly advanced in many real-world applications~\cite{seg_long2015fcn,seg_ronneberger2015unet,seg_chen2017deeplab,seg_zhao2017pyramid,seg_chen2018deeplabv3}.
Most recently, Vision Transformers (ViTs)~\cite{dosovitskiy2021vit,vitn_liu2021swin,vit_wang2021pvt,vitn_chen2021crossvit,vitn_wu2021cvt,vitn_yuan2021t2tvit,vitn_han2021transformer,vitn_yu2022metaformer,vitn_cai2023efficientvit} have remarkably reformed this field.
Concretely, SETR~\cite{vit_setr} adopts ViT as a backbone to extract features, achieving impressive performance. PVT~\cite{vit_wang2021pvt} is the first work to introduce a pyramid structure in Transformer. SegFormer~\cite{vit_xie2021segformer}, with hierarchical MiT encoder and a lightweight All-MLP decoder, serves as a solid baseline for semantic segmentation. 
These ViT-based segmentation approaches demonstrate the potential of a pure Transformer framework in dense prediction tasks.
However, when it comes to semi-supervised setting, it is hard for such a framework to achieve the same advances. The reason is that these Transformer-based methods have weaker inductive bias compared with CNN-based methods and heavily rely on large amount of training data~\cite{dosovitskiy2021vit}. Thus, appropriate designs are desirable to train these methods in low-data regime.

\subsection{Semi-supervised Semantic Segmentation}

Conventional semi-supervised semantic segmentation (SSSS) methods focus on how to better leverage the unlabeled data with framework level designs, and use basic C\textbf{NNs-based} segmentation model as backbone. 
Two most remarkable basic frameworks are the teacher-student~\cite{MT,U2PL,GTASeg,AugSeg,DGCL} and CPS (Cross Pseudo Supervision)~\cite{CPS,UniMatch,CCVC}.
Specifically, in the teacher-student-based framework, a student model is first trained with labeled data, then a teacher model updated by the exponential moving averaging (EMA) of the student is used to generate the pseudo labels. In the CPS-based framework, the labeled data are concatenated with the unlabeled data within a mini-batch, and the model optimize the two flows simultaneously with consistency regularization.

Recently, several works have explored \textbf{Transformer-based} models to extend the border of SSSS. Specifically, SemiCVT~\cite{SemiCVT} presents an inter-model class-wise consistency to complement the class-level statistics of CNNs and Transformer in a cross-teaching manner.
Other methods~\cite{DiverseCotraining,DLG} utilize ViTs as a diverse model to CNNs in CPS-based framework~\cite{CPS}.
These methods can be summarized as CNN-Transformer-based methods and the way they utilize transformer are naive, hence restrict the full potential of ViT. Unlike these methods, we shed new light on a pure-transformer SSSS method with architecture level designs.


\section{\methodName{}: Reborn Labeled Features from Unlabeled Features}

\subsection{Baseline and Overview}\label{sec:overview}

General semi-supervised semantic segmentation dataset comprises limited labeled data and plenty of unlabeled data.
Assume that the entire dataset comprises of $N_L$ labeled samples $\{(x_i^l,y_i)\}_{i=1}^{N_L}$ and $N_U$ unlabeled samples $\{x_i^u\}_{i=1}^{N_U}$, where $x_i \in \mathbb{R}^{3 \times H\times W}$ is the input volume and $y_i \in \mathbb{R}^{K\times H\times W}$ is the ground-truth annotation with $K$ classes. 
Pseudo-labeling-base methods train the model with ground truth supervision first, then generate pseudo labels $\hat{y}$ with the pretrained model as supervision to the unlabeled data. The most basic objective function is formulated as:
\begin{equation}
    \label{eq:loss}
    \mathcal{L} = \mathcal{L}_{s} + \mathcal{L}_{u}
    =\frac{1}{N_L}\sum_{i=0}^{N_L}\mathcal{L}_{CE}(p^{l}_i, y_i) + \frac{1}{N_U}\sum_{j=0}^{N_U} \mathcal{L}_{CE}(p^{u}_{j}, \hat{y}_{j})
\end{equation}
where $\mathcal{L}_{CE}$ is the cross-entropy loss. 

Increasing methods have been proposed to extend this formulation, especially for the unsupervised loss, making this formulation and the training pipeline complicated.
In this work, we focus on introducing \textit{architecture-level} modifications to enhance the performance of pseudo-labeling in semi-supervised semantic segmentation.
We start with the basic formulation of pseudo-labeling as our baseline and then propose the \methodName{} module, which is integrated between the encoder and decoder of the ViT-based segmentation model.
The \methodName{} module, as illustrated in Fig.~\ref{fig:framework} and Algorithm~\ref{algo:algo}, consists a Channel-wise Cross-Attention module (\S\ref{sec:cross_attention}), a \SM{} to store unlabeled features for better reconstruction of the labeled features (\S\ref{sec:memory_bank}), and a Channel-wise Semantic Grouping strategy to better update the \SM{} (\S\ref{sec:grouping}).

\subsection{Channel-wise Cross-attention: the Core of \methodName{}}\label{sec:cross_attention}


To alleviate the dominance of the biased labeled data brought by the separate training scheme, we propose \methodName{} to directly intervene in the labeled data training flow with unlabeled data.
Typically, different feature channels encode distinct semantic information. Compared to patch tokens, these channel-wise features encompass richer contextual information that is more general across various input images (refer to Figure~\ref{fig:vis_membank}).
We leverage this contextual information from unlabeled data to reconstruct the features of labeled data as a robust regularization using the channel-wise cross-attention mechanism~\cite{wang2022uctransnet,ding2022davit}. Within this mechanism, labeled data features serve as queries, while unlabeled features serve as keys and values, as depicted in Figure~\ref{fig:framework}.
Specifically, we calculate the similarity between each channel of the labeled features and the unlabeled features, and the unlabeled channels with the highest similarities play a more significant role in reconstructing the labeled features.
The underlying rationale is that even though the unlabeled features may originate from different classes compared to the labeled features, they still possess generic information with channel wise that can be shared, such as textures.

Given the hidden features $[h^l, h^u]$ of labeled and unlabeled data after the encoder stage, we split them then set $h^l$ as \textit{query} and $h^u$ as \textit{key} and \textit{value} in multi-head manner: 
\begin{equation}
    q=h^l w_{q},k=h^u w_{k},v=h^u w_{v}
\label{eq2}
\end{equation}
where $w_{q}, w_{k}, w_{v}\in\mathbb{R}^{C \times 2C}$ are transformation weights, $h^l, h^u \in \mathbb{R}^{C \times d}$, $d$ is the sequence length (number of patches) and $C$ is the channel dimension.
The channel-wise attention is defined as:
\begin{equation}
\hat{h}^{l} = \mathbf{M} v^\top  = \sigma[\psi(q^\top k)] v^\top w_{out}
\label{eq3}
\end{equation}
where $\psi(\cdot)$ and $\sigma(\cdot)$ denote the instance normalization~\citep{IN} and the softmax function, $w_{out} \in \mathbb{R}^{2C \times C}$.
In contrast to the traditional self-attention mechanism, channel-wise attention can capture long-range dependencies among channels.
To further refine the hidden features of the unlabeled data, we also apply a channel-wise self-attention. The formulation to obtain the refined unlabeled feature $\hat{h}^{u}_i$ is similar to Eq.~\ref{eq2} \& \ref{eq3}, with the only difference being that $q$ is replaced with $q=h^u w_q$.
The reconstructed $\hat{h}^{l}_i$ and $\hat{h}^{u}_i$ are subsequently fed into the decoder to generate the final predictions.


\begin{figure}[!t]
    \centering
    \includegraphics[width=\linewidth]{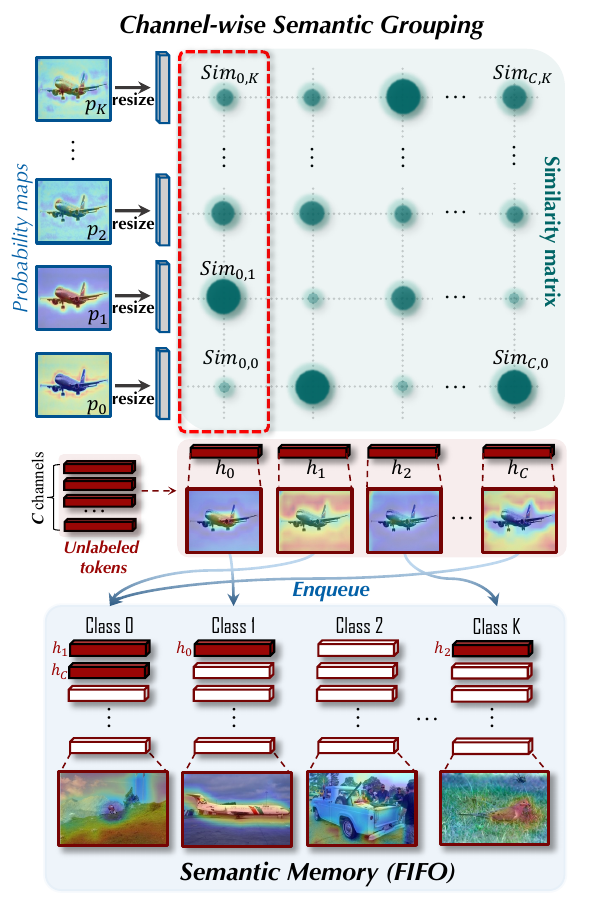}
    \caption{Illustration of the Class-wise Semantic Memory Bank (\S\ref{sec:memory_bank}) and the Channel-wise Semantic Grouping (\S\ref{sec:grouping}). $Sim_{i,j}$ denotes the similarity between $i^{th}$  channel of the unlabeled hidden feature and the $j^{th}$ probability map. The dash lines give some visual examples of some channels. Take the column with \textcolor{red}{red box} as an example, $h_0$ has the largest similarity  with $p_1$ ($Sim_{0,1}$), so it should be added to Class 1 slot of the Semantic Memory.}
    \label{fig:memory_bank}
\end{figure}

\subsection{Semantic Memory for Enlarging the Feature Space of \methodName{}}\label{sec:memory_bank}
Directly utilizing the unlabeled features within a single mini-batch is not enough to effectively reborn the labeled features.
To overcome this limitation, we need to expand the unlabeled feature space.
To achieve this, we introduce a first-in-first-out queue, Class-wise Semantic Memory (S-Mem), to store a significant number of unlabeled features, as shown in Figure~\ref{fig:memory_bank} bottom. This memory allows us to efficiently access a broader range of unlabeled features for the reconstruction process.

S-Mem has a shape of $\mathbb{R}^{K \times C \times d}$, where $K$ represents the number of classes. For each class, S-Mem stores $C$ channels, and each channel consists of $d$ patches.
During the training process, we utilize the semantic memory to replace the original \textit{key} and \textit{value} components of the unlabeled features ($k^T$ and $v^T$ in Eq.~\ref{eq3}).
In the next sub-section \S\ref{sec:grouping}, we demonstrate how to update each class slot with accurate channels containing class-specific semantic information.

\definecolor{commentcolor}{RGB}{110,154,155}   
\newcommand{\PyComment}[1]{\ttfamily\textcolor{commentcolor}{\# #1}}  
\newcommand{\PyCode}[1]{\ttfamily\textcolor{black}{#1}} 

\begin{algorithm}[t]
\scriptsize
\SetAlgoLined
\PyComment{Take Pascal dataset as an example, where the crop size is $513\times513$} \\
\PyComment{channel num = 512, token num = 289, class num=21} \\
\hspace*{\fill} \\

\PyComment{p\_u$\in \mathbb{R}^{289 \times 21}$: probability token} \\
\PyCode{p\_u = flatten(decoder.forward(h\_u).resize(17, 17))} \\
\hspace*{\fill} \\

\PyComment{h\_l$\in \mathbb{R}^{289 \times 512}$: labeled hidden feature map} \\
\PyComment{h\_u$\in \mathbb{R}^{289 \times 512}$: unlabeled hidden feature map} \\
\PyCode{def cross\_attn((h\_l,y), h\_u, p\_u):} \\
\Indp   

    \PyComment{channel2class similarity: 21 $\times$ 512} \\
    \PyCode{sim\_prob\_chl = mm(p\_u.T, h\_u)} \PyComment{see eq (4)} \\
    \PyComment{channel pseudo label: 1 $\times$ 512} \\
    \PyCode{chl\_pseu = argmax(sim\_prob\_chl, dim=0)} \\

    \PyCode{for cls in range(21):} \\
        \Indp
        \PyComment{group channels belonging to class cls} \\
        \PyCode{index = nonzero(where(chl\_pseu==cls, 1, 0))} \\
        \PyCode{chls = h\_u(:, index)} \\
        \PyCode{N\_u = len(index)} \\
        
        \PyComment{enqueue channels into semantic memory} \\
        \PyCode{enqueue(S-Mem, chls, cls)} \\
        \PyCode{dequeue(S-Mem, cls)} \\
        \Indm
    
    \PyComment{queries: 289 $\times$ 512} \\
    \PyCode{q = l\_q.forward(h\_l)} \\
    \PyComment{Keys: 289 $\times$ 512} \\
    \PyCode{k = l\_k.forward(flatten(S-Mem(:,:,:N\_u))} \\
    \PyComment{values: 289 $\times$ 512} \\
    \PyCode{v = l\_v.forward(flatten(S-Mem(:,:,N\_u:))} \\
    \PyComment{channel similarity: 512 $\times$ 512} \\
    \PyCode{m = mm(q.T, k)} \\
    \PyComment{reborn labeled feature: 289 $\times$ 512} \\
    \PyCode{q\_re = mm(m, v.T).T} \\
    \PyCode{return q\_re} \\ 
\Indm
\caption{PyTorch-style pseudocode for \methodName{} module.}
\label{algo:algo}
\end{algorithm}

\subsection{Channel-wise Semantic Grouping}\label{sec:grouping}

Storing previous unlabeled features in a naive manner is not suitable for semantic segmentation tasks, as classes are often imbalanced. 
To ensure sufficient semantic information for each class, it is necessary to establish a class-balanced semantic memory.
Thus, we introduce a Channel-wise Semantic Grouping strategy to determine the semantic representation of each channel in the unlabeled feature, and then group them and add into the corresponding class slot of S-Mem, as illustrated in Figure~\ref{fig:memory_bank}.
Specifically, we calculate the similarity between the unlabeled feature $h^u \in \mathbb{R}^{C\times d}$ and the probability token $\hat{p} \in \mathbb{R}^{K \times d}$. The probability token is resized and reshaped vector obtained from the probability map $p\in \mathbb{R}^{K \times H \times W}$, which is the soft prediction of the segmentation network and contains the overall semantic information of $h^u$. Once the semantic representation is determined for each channel, we can store them in the memory according to their respective semantic categories.
The similarity matrix $Sim \in \mathbb{R}^{K\times C}$ is defined as:
$Sim_{i,j} = \psi(\hat{p}_{j}^\top h^u_i)$.
In this way, by comparing the semantic information of $i^{th}, i\in [0,C]$ channel and the overall semantic context of the image, we could determine the most likely semantic category that the $i^{th}$ channel represents. 
Then, we group together the channels that represent the same semantic category and enqueue them into the appropriate class slot within the semantic memory.

\textbf{In the inference stage}, although we can still utilize the cross-attention mechanism with S-Mem, it often comes with a computational burden and provides only marginal improvements in performance. To enhance efficiency, we opt to remove S-Mem and CSG in the inference stage. As a result, the two inputs of cross-attention become identical, which can be simplified to self-attention.

\section{Experiments}
\subsection{Experimental Setup}
\paragraph{Datasets.} We evaluate our \methodName{} on three widely-used datasets:
\textbf{PASCAL VOC 2012}~\cite{everingham2015pascal} is a dataset consisting of around 4,000 samples. This dataset is partitioned into three subsets: train, validation, and test. The train set comprises 1,464 images, the validation set contains 1,449 images, and the test set consists of 1,456 images. This dataset includes pixel-level annotations for a total of 21 categories, including the background. In accordance with the conventions established in ~\cite{CPS,U2PL}, an additional set of 9,118 coarsely labeled images from the SBD dataset~\cite{hariharan2011SBD} is incorporated as a complement to the training data, which is referred to as the \textit{augmented} set.
The \textbf{Cityscapes} dataset~\cite{cordts2016cityscapes}, is specifically designed for the purpose of urban scene understanding. It comprises a collection of 5,000 images that have been meticulously annotated with fine-grained pixel-level details. These images are divided into three sets: train, validation, and test. The train set contains 2,975 images, the validation set consists of 500 images, and the test set includes 1,524 images. In the Cityscapes dataset, there are 19 semantic categories that are organized into 6 super-classes.
\textbf{COCO}~\cite{lin2014coco} stands out for its extensive annotations and wide variety of object categories. The dataset includes 118,000 training images and 5,000 validation images. It covers 80 object categories and encompasses both indoor and outdoor scenes.

\begin{table}[!t]
    \centering
    \resizebox*{\linewidth}{!}{
    \begin{tabular}{l|ccccc}
    \toprule
        \rowcolor{cyan!20} \textbf{PASCAL} \textit{original} & 1/16 (92) & 1/8 (183) & 1/4 (366) & 1/2 (732) & Full(1464) \\ 
        \midrule
        SupBaseline-CNN & 45.1 & 55.3 & 64.8 & 69.7 & 73.5 \\ 
        SupBaseline-ViT & 50.65 & 63.62 & 70.76 & 75.44 & 77.01 \\ 
        \midrule
        CutMixSeg~\cite{CutMixSeg} \scriptsize \textcolor{gray}{[BMVC'20]} & 55.58 & 63.20 & 68.36 & 69.84 & 76.54 \\ 
        PseudoSeg~\cite{PseudoSeg} \scriptsize \textcolor{gray}{[ICLR'21]} & 57.60 & 65.50 & 69.14 & 72.41 & ~ \\ 
        CPS~\cite{CPS} \scriptsize \textcolor{gray}{[CVPR'21]} & 64.07 & 67.42 & 71.71 & 75.88 & - \\ 
        PC2Seg~\cite{PC2Seg} \scriptsize \textcolor{gray}{[ICCV'21]} & 57.00 & 66.28 & 69.78 & 73.05 & 74.15 \\ 
        PS-MT~\cite{PSMT} \scriptsize \textcolor{gray}{[CVPR'22]} & 65.80 & 69.58 & 76.57 & 78.42 & 80.01 \\ 
        ST++~\cite{ST++} \scriptsize \textcolor{gray}{[CVPR'22]} & 65.20 & 71.00 & 74.60 & 77.30 & 79.10 \\ 
        U$^2$PL~\cite{U2PL} \scriptsize \textcolor{gray}{[CVPR'22]} & 67.98 & 69.15 & 73.66 & 76.16 & 79.49 \\ 
        GTA-Seg~\cite{GTASeg} \scriptsize \textcolor{gray}{[NeurIPS'22]} & 70.02 & 73.16 & 75.57 & 78.37 & 80.47 \\ 
        SemiCVT~\cite{SemiCVT} \scriptsize \textcolor{gray}{[CVPR'23]} & 68.56 & 71.26 & 74.99 & 78.54 & 80.32 \\ 
        FPL~\cite{FPL} \scriptsize \textcolor{gray}{[CVPR'23]} & 69.30 & 71.72 & 75.73 & 78.95 & - \\ 
        CCVC~\cite{CCVC} \scriptsize \textcolor{gray}{[CVPR'23]} & 70.2 & 74.4 & 77.4 & 79.1 & 80.5 \\ 
        AugSeg~\cite{AugSeg} \scriptsize \textcolor{gray}{[CVPR'23]} & 71.09 & 75.45 & 78.8 & \underline{80.33} & 81.36 \\ 
        DGCL~\cite{DGCL} \scriptsize \textcolor{gray}{[CVPR'23]} & 70.47 & 77.14 & 78.73 & 79.23 & 81.55 \\ 
        UniMatch~\cite{UniMatch} \scriptsize \textcolor{gray}{[CVPR'23]} & \underline{75.2} & \underline{77.19} & \underline{78.8} & 79.9 & - \\ 
        ESL~\cite{ESL} \scriptsize \textcolor{gray}{[ICCV'23]} & 70.97 & 74.06 & 78.14 & 79.53 & \underline{81.77} \\ 
        LogicDiag~\cite{LogicDiag} \scriptsize \textcolor{gray}{[ICCV'23]} & 73.25 & 76.66 & 77.93 & 79.39 & - \\ 
        \midrule
        \rowcolor{red!10} \textbf{AllSpark (Ours)} & \textbf{76.07}  & \textbf{78.41}  & \textbf{79.77}  & \textbf{80.75}  & \textbf{82.12}  \\ \bottomrule
    \end{tabular}}
    \caption{Comparison with SOTAs on \textbf{Pascal} \textit{original} dataset.}
    \label{tab:sota_pascal_org}
\end{table}

\begin{table}[!t]
    \centering
    \resizebox*{\linewidth}{!}{
    \begin{tabular}{l|cccc}
    \toprule
        \rowcolor{cyan!20} \textbf{Cityscapes} & 1/16 (186) & 1/8 (372) & 1/4 (744) & 1/2 (1488) \\  
        \midrule 
        SupBaseline-CNN & 66.3 & 72.8 & 75.0 & 78.0 \\ 
        SupBaseline-ViT & 65.69  & 70.52  & 74.01  & 78.14 \\ \midrule
        CPS~\cite{CPS} \scriptsize \textcolor{gray}{[CVPR'21]} & 69.78 & 74.31 & 74.58 & 76.81 \\ 
        AEL~\cite{AEL} \scriptsize \textcolor{gray}{[NeurIPS'21]} & 74.45 & 75.55 & 77.48 & 79.01 \\ 
        ST++~\cite{ST++} \scriptsize \textcolor{gray}{[CVPR'22]} & 67.64 &73.43 &74.64 &77.78 \\ 
        
        U$^2$PL~\cite{U2PL} \scriptsize \textcolor{gray}{[CVPR'22]} & 74.90 &76.48 &78.51 &79.12 \\ 
        
        GTA-Seg~\cite{GTASeg} \scriptsize \textcolor{gray}{[NeurIPS'22]}  & 69.38 & 72.02 & 76.08 & - \\ 
        
        SemiCVT~\cite{SemiCVT} \scriptsize \textcolor{gray}{[CVPR'23]} & 72.19 &75.41 &77.17 &79.55 \\ 
        
        FPL~\cite{FPL} \scriptsize \textcolor{gray}{[CVPR'23]} & 75.74 &78.47 &79.19  & - \\ 
        
        CCVC~\cite{CCVC} \scriptsize \textcolor{gray}{[CVPR'23]} & 74.9 & 76.4 & 77.3 & - \\ 
        AugSeg~\cite{AugSeg} \scriptsize \textcolor{gray}{[CVPR'23]} & 75.22 &77.82 &79.56 &80.43 \\ 
        DGCL~\cite{DGCL} \scriptsize \textcolor{gray}{[CVPR'23]} & 73.18 & 77.29 & 78.48 & 80.71 \\ 
        UniMatch~\cite{UniMatch} \scriptsize \textcolor{gray}{[CVPR'23]} & 76.6 &77.9 & 79.2 & 79.5 \\ 
        ESL~\cite{ESL} \scriptsize \textcolor{gray}{[ICCV'23]} & 75.12 & 77.15 & 78.93 & 80.46 \\ 
        LogicDiag~\cite{LogicDiag} \scriptsize \textcolor{gray}{[ICCV'23]} & 76.83 & 78.90 & \underline{80.21} & \underline{81.25} \\ 
        CFCG~\cite{CFCG} \scriptsize \textcolor{gray}{[ICCV'23]} & \underline{77.28} &\underline{79.09} &80.07 &80.59 \\ 
        \midrule
        \rowcolor{red!10} \textbf{AllSpark (Ours)} & \textbf{78.33} & \textbf{79.24} & \textbf{80.56}& \textbf{81.39} \\ \bottomrule
    \end{tabular}}
    \caption{Comparison with SOTAs on \textbf{Cityscapes} dataset.}
    \label{tab:sota_city}
\end{table}

\noindent\textbf{Implementation Details.}
For the three datasets, we all utilized SGD as the optimizer, and the poly scheduling to adjust the learning rate as $lr = lr_{init} \cdot (1 - \frac{i} {I} )^{0.9} $, where $lr_{init}$ represents the initial learning rate, $i$ is the current iteration, and $I$ is the max number of iterations. 
For the \textit{PASCAL VOC 2012 dataset}, we set the initial learning rate as 0.001. Please note that the learning rate for the AllSpark and the decoder is 5 times of the backbone network. We train the model for 80 epochs. For the \textit{Cityscapes dataset}, the initial learning rate is set as 0.005, and we train the model for 240 epochs. It is worth noting that for the Cityscapes dataset, we employ two approaches used in previous works~\cite{UniMatch,U2PL,CPS,AEL,PSMT}, i.e., online hard example mining (OHEM) and sliding window evaluation. For the \textit{COCO dataset}, we set the initial learning rate of 0.001, and train the model for 10 epochs.

\noindent\textbf{Baselines.}
We report two supervised baselines among different ratios of labeled data: a CNN-based model, RestNet-101~\cite{resnet} with DeepLabV3+~\cite{seg_chen2018deeplabv3}, referred to as \textbf{SupBaseline-CNN}, and a ViT-based model, SegFormer-B5~\cite{vit_xie2021segformer}, referred to as \textbf{SupBaseline-ViT}. Additionally, we employ a simple pseudo-labeling scheme outlined in section \S\ref{sec:overview} with SegFormer-B5 as our semi-supervised baseline, denoted as \textbf{SemiBaseline}.

\begin{table}[!t]
    \centering
    \resizebox*{\linewidth}{!}{
    \begin{tabular}{l|cccc}
    \toprule
        \rowcolor{cyan!20} \textbf{PASCAL} \textit{augmented} & 1/16 (662) & 1/8 (1323) & 1/4 (2646) & 1/2 (5291) \\ 
        \midrule
        SupBaseline-CNN & 67.5 & 71.1 & 74.2 & - \\ 
        SupBaseline-ViT & 72.01 & 73.20 & 76.62 & 77.61 \\ 
        \hline
        CutMixSeg~\cite{CutMixSeg} \scriptsize \textcolor{gray}{[BMVC'20]} & 72.56 & 72.69 & 74.25 & 75.89 \\ 
        PseudoSeg~\cite{PseudoSeg} \scriptsize \textcolor{gray}{[ICLR'21]} & - & - & - & - \\ 
        AEL~\cite{AEL} \scriptsize \textcolor{gray}{[NeurIPS'21]} & 77.20 & 77.57 & 78.06 & 80.29 \\ 
        CPS~\cite{CPS} \scriptsize \textcolor{gray}{[CVPR'21]} & 72.18 & 75.83 & 77.55 & 78.64 \\ 
        PC2Seg~\cite{PC2Seg} \scriptsize \textcolor{gray}{[ICCV'21]} & - & - & - & - \\ 
        PS-MT~\cite{PSMT} \scriptsize \textcolor{gray}{[CVPR'22]} & 75.50 & 78.20 & 78.72 & 79.76 \\ 
        ST++~\cite{ST++} \scriptsize \textcolor{gray}{[CVPR'22]} & 74.70 & 77.90 & 77.90 & - \\ 
        FPL~\cite{FPL} \scriptsize \textcolor{gray}{[CVPR'23]} & 74.98 & 77.75 & 78.30 & - \\ 
        CCVC~\cite{CCVC} \scriptsize \textcolor{gray}{[CVPR'23]} & 77.2 & 78.4 & 79.0 & - \\ 
        AugSeg~\cite{AugSeg} \scriptsize \textcolor{gray}{[CVPR'23]} & 77.01 & 78.20 & 78.82 & - \\ 
        DGCL~\cite{DGCL} \scriptsize \textcolor{gray}{[CVPR'23]} & 76.61 & 78.37 & 79.31 & \underline{80.96} \\ 
        UniMatch~\cite{UniMatch} \scriptsize \textcolor{gray}{[CVPR'23]} & 78.1 & 78.4 & 79.2 & - \\ 
        ESL~\cite{ESL} \scriptsize \textcolor{gray}{[ICCV'23]} & 76.36 & 78.57 & 79.02 & 79.98 \\ 
        
        CFCG~\cite{CFCG} \scriptsize \textcolor{gray}{[ICCV'23]} & 76.82 & 79.10 & \underline{79.96} & 80.18 \\ 
        DLG~\cite{DLG} \scriptsize \textcolor{gray}{[ICCV'23]} & \underline{77.75} & \underline{79.31} & 79.14 & 79.54 \\ 
        \midrule
        \rowcolor{red!10} \textbf{AllSpark (Ours)} & \textbf{78.32}  & \textbf{79.98}  & \textbf{80.42}  & \textbf{81.14}  \\ \midrule \midrule
        U$^2$PL$^\dag$~\cite{U2PL} \scriptsize \textcolor{gray}{[CVPR'22]} & 77.21 & 79.01 & 79.30 & 80.50 \\ 
        GTA-Seg$^\dag$~\cite{GTASeg} \scriptsize \textcolor{gray}{[NeurIPS'22]} & 77.82 & 80.47 & 80.57 & \underline{81.01} \\ 
        SemiCVT$^\dag$~\cite{SemiCVT} \scriptsize \textcolor{gray}{[CVPR'23]} & 78.20 & 79.95 & 80.20 & 80.92 \\ 
        UniMatch$^\dag$~\cite{UniMatch} \scriptsize \textcolor{gray}{[CVPR'23]} & \underline{80.94} & \underline{81.92} & 80.41 & - \\ 
        LogicDiag$^\dag$~\cite{LogicDiag} \scriptsize \textcolor{gray}{[ICCV'23]} & 79.65 & 80.24 & \underline{80.62} & 81.00 \\ 
        \midrule
        \rowcolor{red!10}\textbf{AllSpark$^\dag$ (Ours) } & \textbf{81.56}  & \textbf{82.04}  & \textbf{80.92}  & \textbf{81.13}  \\
        
        \bottomrule
    \end{tabular}}
    \caption{Comparison with SOTAs on the \textbf{Pascal} \textit{augmented} dataset. $\dag$ means using the same split as U$^2$PL~\cite{U2PL}.}
    \label{tab:sota_pascal_aug}
\end{table}

\begin{table}[!t]
    \centering
    \resizebox*{\linewidth}{!}{
    \begin{tabular}{l|cccc}
    \toprule
        \rowcolor{cyan!20} \textbf{COCO} & 1/512 (232) & 1/256 (463) & 1/128 (925) & 1/64 (1849) \\  
        \midrule 

        SupBaseline-CNN &22.94  &27.96 &33.60 &37.80 \\

        SupBaseline-ViT &19.66  &26.72 &35.90 &40.89 \\ \hline

        PseudoSeg~\cite{PseudoSeg} \scriptsize \textcolor{gray}{[ICLR'21]} & 29.78 & 37.11 & 39.11 & 41.75 \\ 

        PC2Seg~\cite{PC2Seg} \scriptsize \textcolor{gray}{[ICCV'21]} & 29.94 & 37.53 & 40.12 & 43.67 \\

        MKD~\cite{MKD} \scriptsize \textcolor{gray}{[ACM MM'23]} & 30.24 &38.04 &42.32  & 45.50 \\ 

        UniMatch~\cite{UniMatch} \scriptsize \textcolor{gray}{[CVPR'23]} & {31.86} & {38.88} & {44.35} & {48.17} \\ 

        LogicDiag~\cite{LogicDiag} \scriptsize \textcolor{gray}{[ICCV'23]} &\underline{33.07}  &\underline{40.28} &\underline{45.35} &\underline{48.83} \\ 
        
        \midrule

        \rowcolor{red!10} \textbf{AllSpark (Ours)} & \textbf{34.10} & \textbf{41.65} & \textbf{45.48}& \textbf{49.56} \\ \bottomrule
    \end{tabular}}
    \caption{Comparison with SOTAs on different partitions of the \textbf{COCO} dataset.}
    \label{tab:sota_coco}
\end{table}

\subsection{Comparison with Existing Methods}

\noindent\textbf{PASCAL VOC 2012 original.} Table~\ref{tab:sota_pascal_org} presents a summary of the quantitative comparisons across different label amounts. From this analysis, we make an important observation: our method consistently outperforms the previous state-of-the-art approaches in all scenarios. This highlights the effectiveness of \methodName{} and establishes it as the new state-of-the-art. 

\noindent\textbf{PASCAL VOC 2012 augmented.} Table~\ref{tab:sota_pascal_aug} showcases the comparison results using the augmented training set. Once again, our method establishes new state-of-the-art performance across all partition protocols. It achieves an average gain of 0.47. Furthermore, we conducted a comparison using the same setting as U$^2$PL~\cite{U2PL}, where all labeled data is of high quality. In this scenario, \methodName{} demonstrates an even greater performance gain.

\noindent\textbf{Cityscapes.} Table~\ref{tab:sota_city} quantitatively compares our method against the competitors on Cityscapes val. In spite of the presence of complex street scenes, our method still delivers a solid overtaking trend across different partitions. 

\noindent\textbf{COCO.} In Table~\ref{tab:sota_coco}, we showcase the model performance on the COCO validation set. Notably, leveraging the extensive semantic hierarchy present in COCO, \methodName{} achieves significant performance gains over the leading method in all partitions. These experimental results further validate the effectiveness of \methodName{}.
\begin{table}[ht]
    \centering
    \resizebox*{0.73\linewidth}{!}{
    \begin{tabular}{cccc|cc}
    \toprule
     \multirow{2}{*}{SemiBaseline}  & \multicolumn{3}{c|}{\methodName{}}
      & 1/8 & 1/2 \\
     & CCA & S-Mem & CSG & (183) &  (732)\\\midrule
     \checkmark &  &  &  & 68.71 & 74.40 \\
     \checkmark & \checkmark &  &  & 76.77  & 78.93 \\
     \checkmark & \checkmark & \checkmark  & & 77.62  & 79.71 \\
     \checkmark & \checkmark & \checkmark & \checkmark & \textbf{78.41} & \textbf{80.75}\\ 

    \bottomrule
    \end{tabular}}
    \caption{Effectiveness of the proposed components on \textbf{PASCAL} dataset. We leverage a basic pseudo-labeling scheme (\S\ref{sec:overview}) with the SegFormer~\cite{vit_xie2021segformer} as our baseline, \methodName{} consists of 3 hierarchy components: CCA (Channel-wise Cross-Attention), S-Mem (\SM), and CSG (\CSG{}).}
    \label{tab:components}
\end{table}

\begin{figure}[!t]
    \centering
    \includegraphics[width=0.9\linewidth]{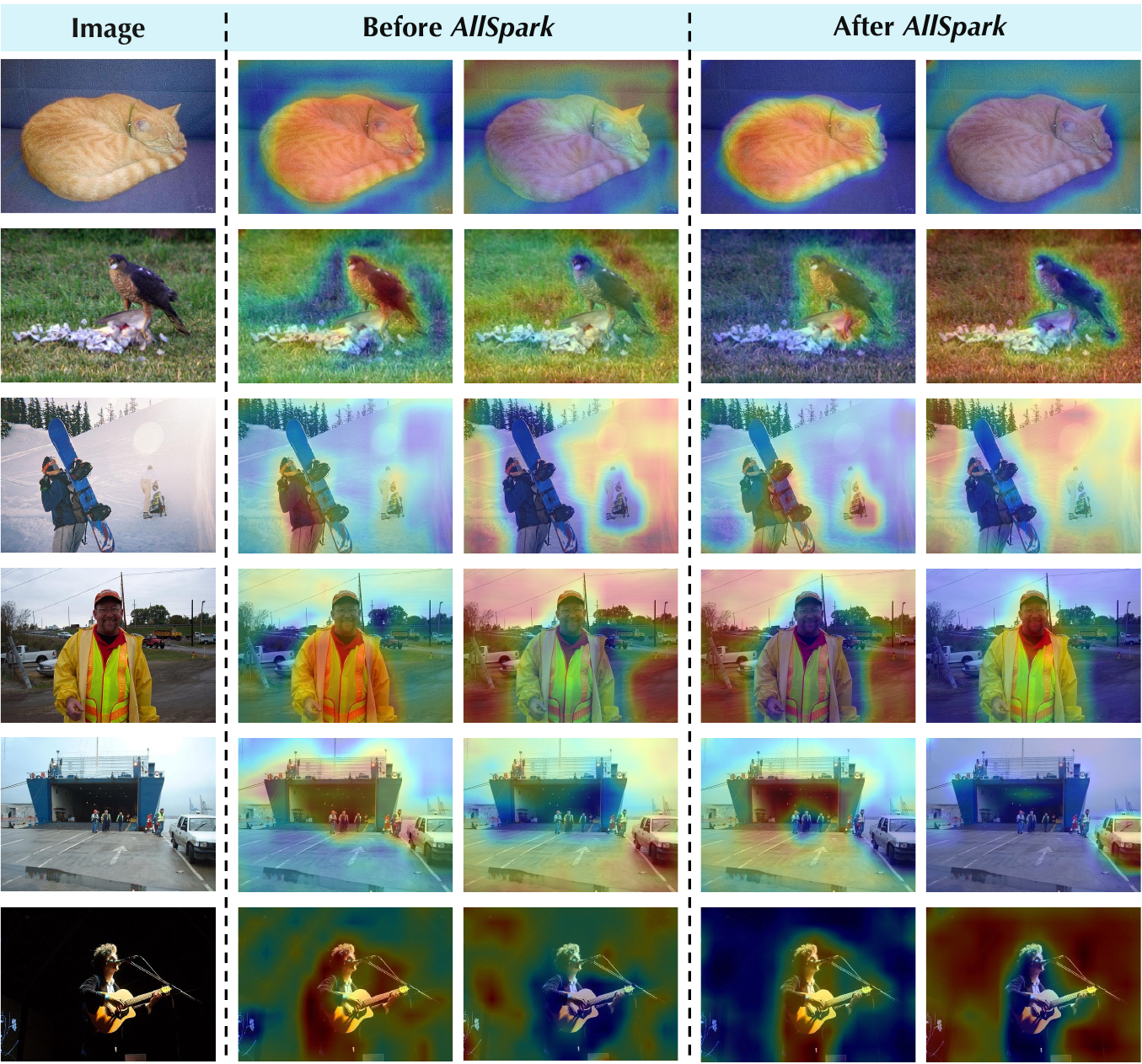}
    \caption{Visualization of labeled feature channels before and after the \methodName{} with the same indexes. The features before \textit{AllSpark} focus on more similar regions, while those after \textit{AllSpark} focus on different objects or context.}
    \label{fig:vis_allspark}
\end{figure}

\begin{figure}[!t]
    \centering
    \includegraphics[width=0.9\linewidth]{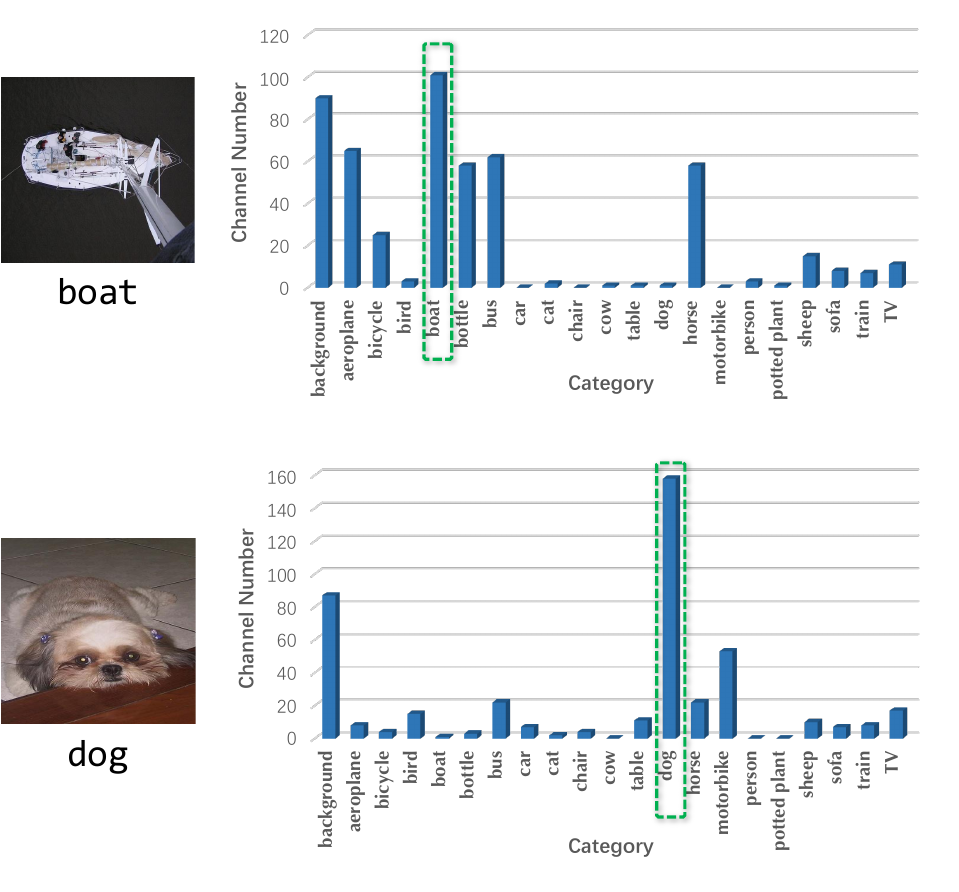}
    \caption{Ablation on the effectiveness of the similarity-guided channel grouping strategy. The y-axis of the bar chart denotes the number of channels to be selected and added to the corresponding slot of the \SM.}
    \label{fig:ablation_grouping}
\end{figure}

\begin{figure*}[!t]
    \centering
    \includegraphics[width=0.83\linewidth]{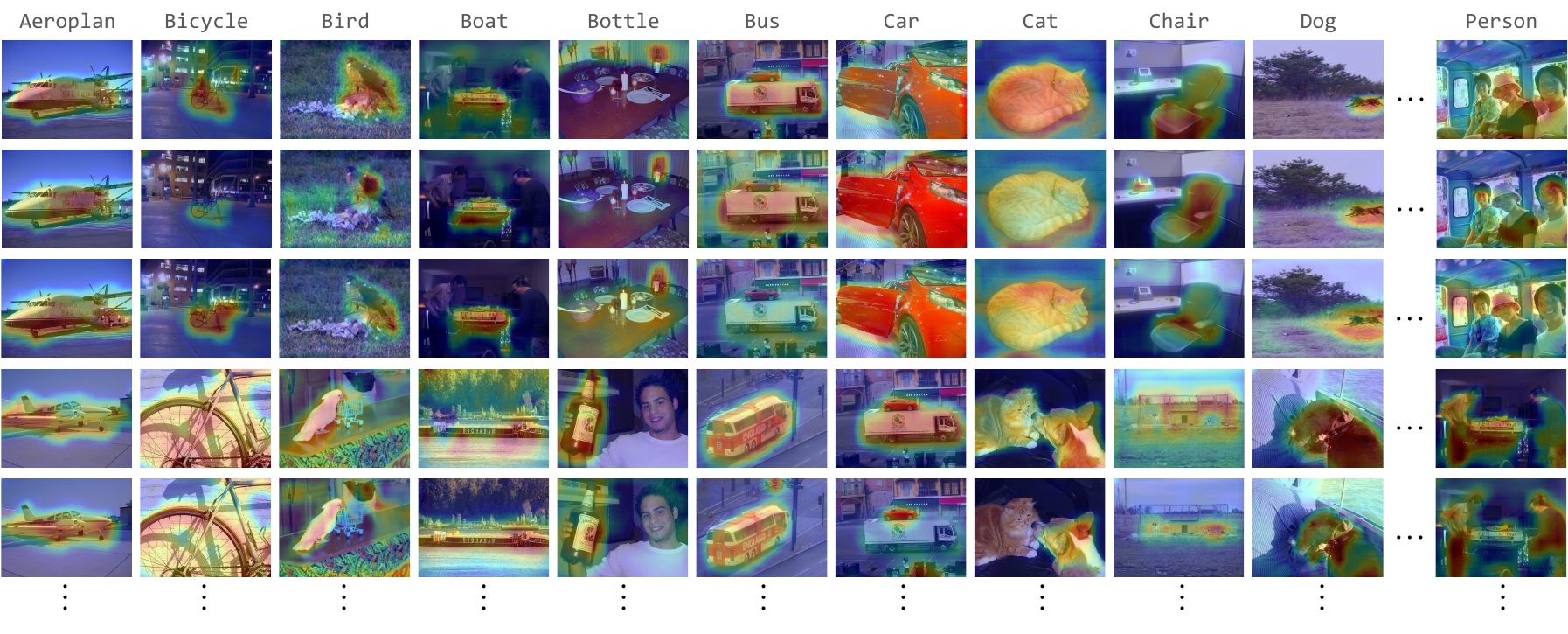}
    \caption{Visualization of part of the the \SM{}. Each image denotes one channel. For better visualization, we run inference on unlabeled training data, so that we can put the corresponding raw image under each channel.}
    \label{fig:vis_membank}
\end{figure*}
\subsection{Ablation Studies}

\noindent\textbf{Effectiveness of the \methodName{} Module.}
We analyze the effectiveness of different components in \methodName{}, \textit{i.e.}, the channel-wise cross attention mechanism (CCA), semantic memory (S-Mem) and channel-wise similarity-guided grouping strategy (CSG) as in Table~\ref{tab:components}.
The three component are hierarchy, \textit{i.e.}, S-Mem is built upon CCA, the core of \methodName{}, and CSG strategy is built for the update of S-Mem.
According to the results, the core of \methodName{}, channel-wise cross-attention mechanism (the second row), helps our method outperform the previous methods, pushing the performance about 8.06\%/4.53\% higher than the baseline (the first row). Further, the semantic memory brings about 0.85\%/0.78\% performance improvements. With all of these components, our method outperforms the baseline by over 9.70\%/5.95\% in mIoU. 
Furthermore, we also provide visual verification of \methodName{} module, as shown in Figure~\ref{fig:vis_allspark}. The reconstructed labeled features after \methodName{} are quite diverse compared with the original features, which is useful for alleviating the dominance of labeled data. Moreover, the features tend to be closer to the target due to the larger feature space compared with a self-attention with the features of itself.

\noindent\textbf{The necessity of the Semantic Memory for enlarging unlabeled feature space.}
As shown in Figure~\ref{fig:vis_membank}, we visualize part of the \SM{}. Overall, the \SM{} stores quite accurate features related to the class slot and indeed enlarges the unlabeled feature space not only in the same classes with the current labeled data, but also the features from other classes as negative samples when calculating the similarity matrix. 

\noindent\textbf{Ablation on the effectiveness of the Channel-wise Semantic Grouping strategy.}
As shown in Table~\ref{tab:components}, compared with blindly adding all channels to the semantic memory, using semantic grouping strategy obtains 0.79\%/0.64\% performance gain. Figure~\ref{fig:ablation_grouping} also verifies that this strategy can assign accurate semantic labels to most channels. Furthermore, the examples in Figure~\ref{fig:ablation_grouping} indicate that the proposed strategy may add similar features from other classes, \textit{e.g.}, some channels of the \verb|boat| are added to some similar classes, \verb|aeroplane|, \verb|bus|, \textit{etc.}, which offers richer features as references to reform other classes. 

\begin{figure}[!t]
    \centering
    \includegraphics[width=\linewidth]{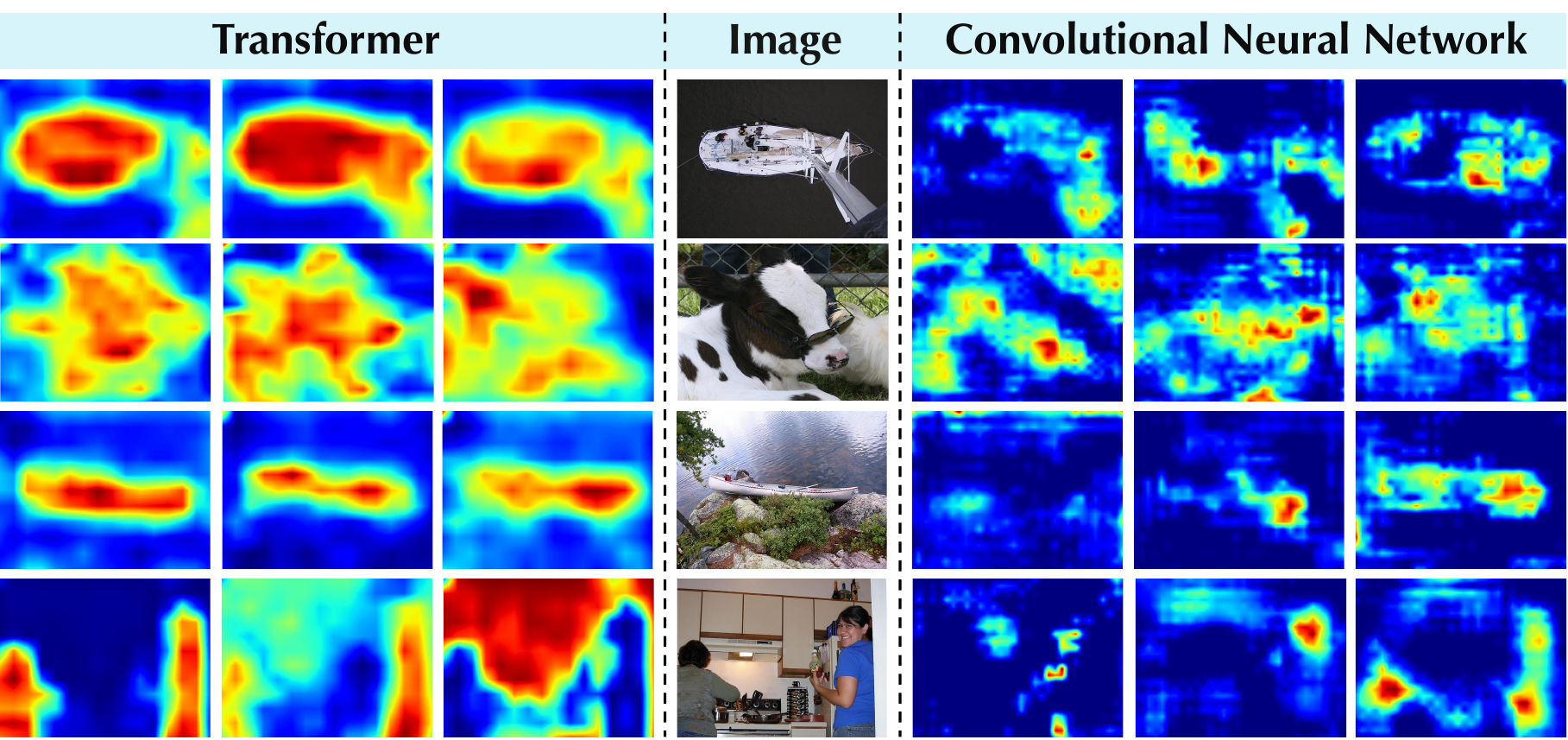}
    \caption{Comparison of Transformer and CNN feature channels.}
    \label{fig:CNNvsViT}
\end{figure}

\noindent\textbf{Why tailored for transformer-based method?}
It might be argued that our improvements are due to a stronger backbone, SegFormer~\cite{vit_xie2021segformer}. 
Considering this concern, we further carry out a comparison with the previous SOTAs with transformer backbones. 
\begin{table}[!t]
    \centering
    \resizebox*{\linewidth}{!}{
\begin{tabular}{@{}l|cccc}
    \toprule
     Backbone  & CCVC~\cite{CCVC} &DGCL~\cite{DGCL} & UniMatch~\cite{UniMatch} & \textbf{Ours} \\ \midrule
     R101+DeepLabV3+  & 74.40 {\ }  & 77.14 {\ }  & 77.19 {\ }  &73.70 {\ } \\
     SegFormer-B4  & 71.07\textcolor{red}{$\downarrow$} & 76.33\textcolor{red}{$\downarrow$} & 76.28\textcolor{red}{$\downarrow$} &77.92\textcolor{teal}{$\uparrow$}\\ 
     SegFormer-B5   & 73.77\textcolor{red}{$\downarrow$} & 76.72\textcolor{red}{$\downarrow$} & 76.56\textcolor{red}{$\downarrow$} & 78.41\textcolor{teal}{$\uparrow$}\\
    \bottomrule
    \end{tabular}
}
    \caption{Ablations on different Segmentation Backbones on \textit{original} PASCAL VOC 2012 dataset - 1/8 (183).}
    \label{tab:backbones}
\end{table}
As shown in Table~\ref{tab:backbones}, when switching to SegFormer-B4/B5, previous methods all have performance drops, since the training of transformer requires more annotated data to avoid over-fitting~\cite{varis2021sequence}.
However, our \methodName{} also exhibits poorer performance with ResNet101 and DeepLabV3+. This can be attributed to the limited receptive fields of CNNs. As shown in Fig.~\ref{fig:CNNvsViT}, when analyzing large objects (\textit{e.g.}, car), the attention maps often prioritize a few salient local parts (\textit{e.g.}, wheels). Consequently, when these features processed by \methodName{}, there is a risk of erroneously querying channels that correspond to similar parts of different objects (\textit{e.g.}, wheels of an aeroplane landing gear), leading to unnecessary noise and performance decay.
\section{Conclusions}
We notice the separate training approach in current SSSS frameworks will cause the dominance of labeled data and thus lead to sub-optimal solutions. 
This separation hinders the effective integration of ViT backbones with SSSS frameworks.
We propose \methodName{} to alleviate this issue by directly reborning the labeled features with the unlabeled features.
To further augment the unlabeled feature space, we introduce a \SM{} (S-Mem) to store previous unlabeled features from different classes.
Then, we can leverage these comprehensive features from all classes to reconstruct more accurate labeled features. To establish the class-balanced S-Mem, we propose \CSG{} strategy, which assigns semantic information to each channel based on its similarity with the probability map.
We present that equipped with the proposed \methodName{}, a naive pseudo-labeling scheme can outperform the SOTAs. 
Finally, \methodName{} improves previous results remarkably on three well-established benchmarks.

\noindent\textbf{Acknowledgement.}
This work is partially supported by the National Natural Science Foundation of China under Grant 62306254, the Hong Kong Innovation and Technology Fund under Grant ITS/030/21, and grants from Foshan HKUST Projects under Grants FSUST21-HKUST10E and FSUST21-HKUST11E.
{
    \small
    \bibliographystyle{ieeenat_fullname}
    \bibliography{main}
}

\clearpage
\setcounter{page}{1}
\maketitlesupplementary
\setcounter{section}{0}
\setcounter{figure}{0}
\setcounter{table}{0}
\renewcommand\thesection{\Alph{section}}
\renewcommand\thetable{\alph{table}}
\renewcommand\thefigure{\alph{figure}}

\noindent\textbf{A. Ablation on Memory Queue Size.} As shown in Table~\ref{tab:queue_size}, we evaluated our method under various queue sizes. The results show that the $1 \times C$ queue size yields the best result. Therefore, we adopted this size.

\begin{figure*}[b]
    \centering    
    \includegraphics[width=\linewidth]{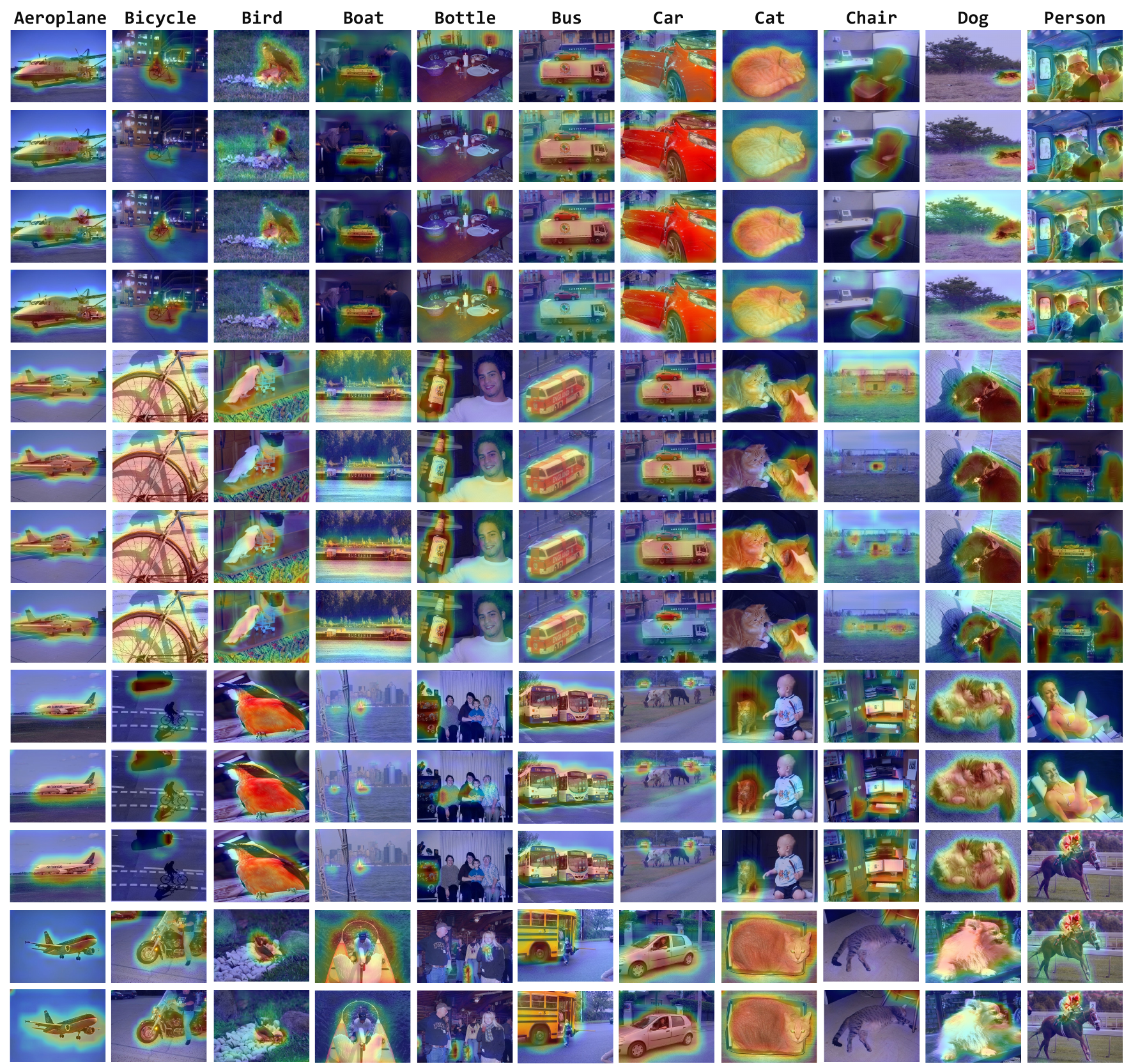}
    \caption{More visualization of the features in Semantic Memory.}
    \label{fig:label1}
\end{figure*}

\begin{figure*}[hbp]
    \centering    
    \includegraphics[width=\linewidth]{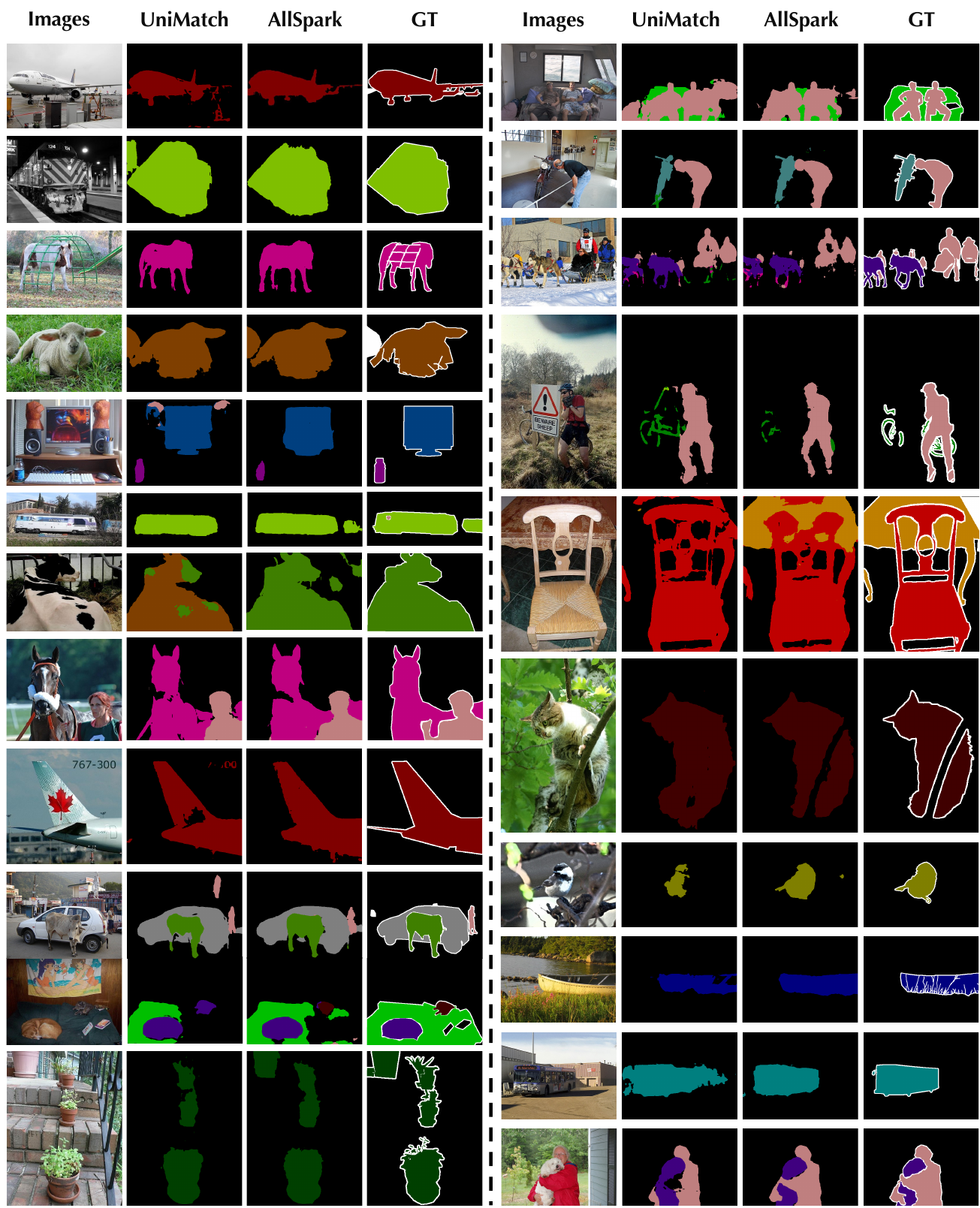}
    \caption{Visualization of the segmentation results on \textbf{Pascal} validation set, compared with UniMatch~\cite{UniMatch}.}
    \label{fig:label2}
\end{figure*}

\noindent\textbf{B. More Visualizations (Figure~\ref{fig:label1} \& ~\ref{fig:label2}).}
\label{sec:rationale}

\newpage

\begin{table}[t]
\centering
\resizebox*{0.85\linewidth}{!}{
    \begin{tabular}{c|cccc}
     Queue Size & $1/2 \times C$ &$1 \times C$ &$2 \times C$ &$4 \times C$\\ \hline
     mIoU  & 77.63 & \textbf{78.41} & 77.98 & OOM
\end{tabular}}
    \caption{$C$: channel number, OOM: out of memory.} 
    \vspace{-0.2cm} 
    \label{tab:queue_size}
\end{table}

\end{document}